\documentclass[letterpaper, 10 pt, conference]{ieeeconf}  

\IEEEoverridecommandlockouts                              

\let\labelindent\relax   

\usepackage{graphicx}
\usepackage[dvipsnames]{xcolor}
\usepackage{subcaption}
\usepackage{cite}
\usepackage{float}
\usepackage{booktabs,ragged2e}
\usepackage{algorithm}
\usepackage{algpseudocode}
\usepackage{amsmath}

\usepackage[flushleft]{threeparttable}

\usepackage[inline]{enumitem}
\setlist{nolistsep}
\newcommand{\il}[1]{\begin{enumerate*}[label=(\roman*)]#1\end{enumerate*}}

\makeatletter\newcommand{\manuallabel}[2]{\def\@currentlabel{#2}\label{#1}}\makeatother

\newcommand*{\schemeref}[1]{Scheme \ref{scheme:#1}} 
\renewcommand{\baselinestretch}{0.850}

\title{\LARGE \bf
Behavior Tree Generation using Large Language Models for Sequential Manipulation Planning with Human Instructions and Feedback
}

\author{Jicong Ao$^{1}$, Yansong Wu$^{1}$, Fan Wu$^{1*}$ and Sami Haddadin$^{1}$
    \thanks{$^{1}$Chair of Robotics and Systems Intelligence, Technical University of Munich, 80992, Munich, Germany
        {\tt\small \{jicong.ao, yansong.wu, f.wu, haddadin\}@tum.de}}%
    \thanks{$^{*}$Corresponding author.}
}

\begin{document}

\maketitle
\thispagestyle{empty}
\pagestyle{empty}

\section{Introduction}

Sequential manipulation planning has been a critical imperative to achieve a higher level of autonomy in robotics. 
Classical approaches to address task planning problems are based on symbolic formalisms, such as Planning Domain Definition Language (PDDL) \cite{McDermott1998PDDL}, and search for state transition plans to reach task goals. 
In practice, such task plans are often programmed as Finite State Machines (FSMs), which incorporate expert knowledge specifying control and execution details. 
Due to its limitation of scalability \cite{iovinoSurveyBehaviorTrees2022}, Behavior trees (BTs), which represent policies in a state-less, hierarchical tree structure, have gained increasing popularity for complex task planning. 
Its advantages of modularity, reusability and reactivity, make it a more desired formalism for long-horizon manipulation tasks. 

Despite being more efficient to program, maintain, and modify than FSMs \cite{iovino2023programming}, manually programming BTs still requires significant effort and is time-consuming. To automate the generation of BT-based task plans in the context of robot manipulation, progress has been made by using \il{\item symbolic planning \cite{colledanchiseBlendedReactivePlanning2019}, \item learning from demonstration \cite{French2019learningBT}, and \item learning by reinforcement \cite{Xu2022BTRL}.} However, cross-domain transfer and robust replanning in dynamic environments still remain challenging. 
Moreover, it is still an open question of how to facilitate intuitive and straightforward human-robot interaction (HRI) and collaboration in BT-based robot control systems.

A new approach to address robot task planning has emerged due to the rapid advance in Large Language Models (LLMs) and Visual-Language Models (VLMs). Recent progress such as ProgPrompt \cite{singhProgPromptProgramGeneration2023} and PaLME \cite{driessPaLMEEmbodiedMultimodal2023} has come from exploiting the capability of semantic understanding of LLMs (VLMs) and leveraging their reasoning capability which can be improved via in-context learning \cite{wangSelfInstructAligningLanguage2023,weiChainofThoughtPromptingElicits,yaoTreeThoughtsDeliberate2023,ning2024skeletonofthought}.

In this work, we propose an LLM-based BT generation framework to leverage the strengths of both for sequential manipulation planning. To enable human-robot collaborative task planning and enhance intuitive robot programming by nonexperts, the framework takes human instructions to initiate the generation of action sequences and human feedback to refine BT generation in runtime. 
The framework is presented in Sec. \ref{sec:framework} by first outlining a basic workflow and then four possible methods for BT generation. All presented methods within the framework are tested on a real robotic assembly example which uses a gear set model from Siemens Robot Assembly Challenge. We use a single manipulator with a tool-changing mechanism, a common practice in flexible manufacturing to facilitate robust grasping of a large variety of objects.
Experimental results are reported in Sec.~\ref{sec:result} which compare the results in terms of success rate, logical coherence, executability, time consumption, and token consumption. 
To our knowledge, this is the first human-guided LLM-based BT generation framework that unifies various plausible ways of using LLMs to fully generate BTs that are executable on the real testbed and take into account granular knowledge of tool use. 


\section{Framework}\label{sec:framework}
\begin{figure*}[tp]
  \centering
  \begin{subfigure}[b]{0.55\textwidth}
    \includegraphics[width=\textwidth]{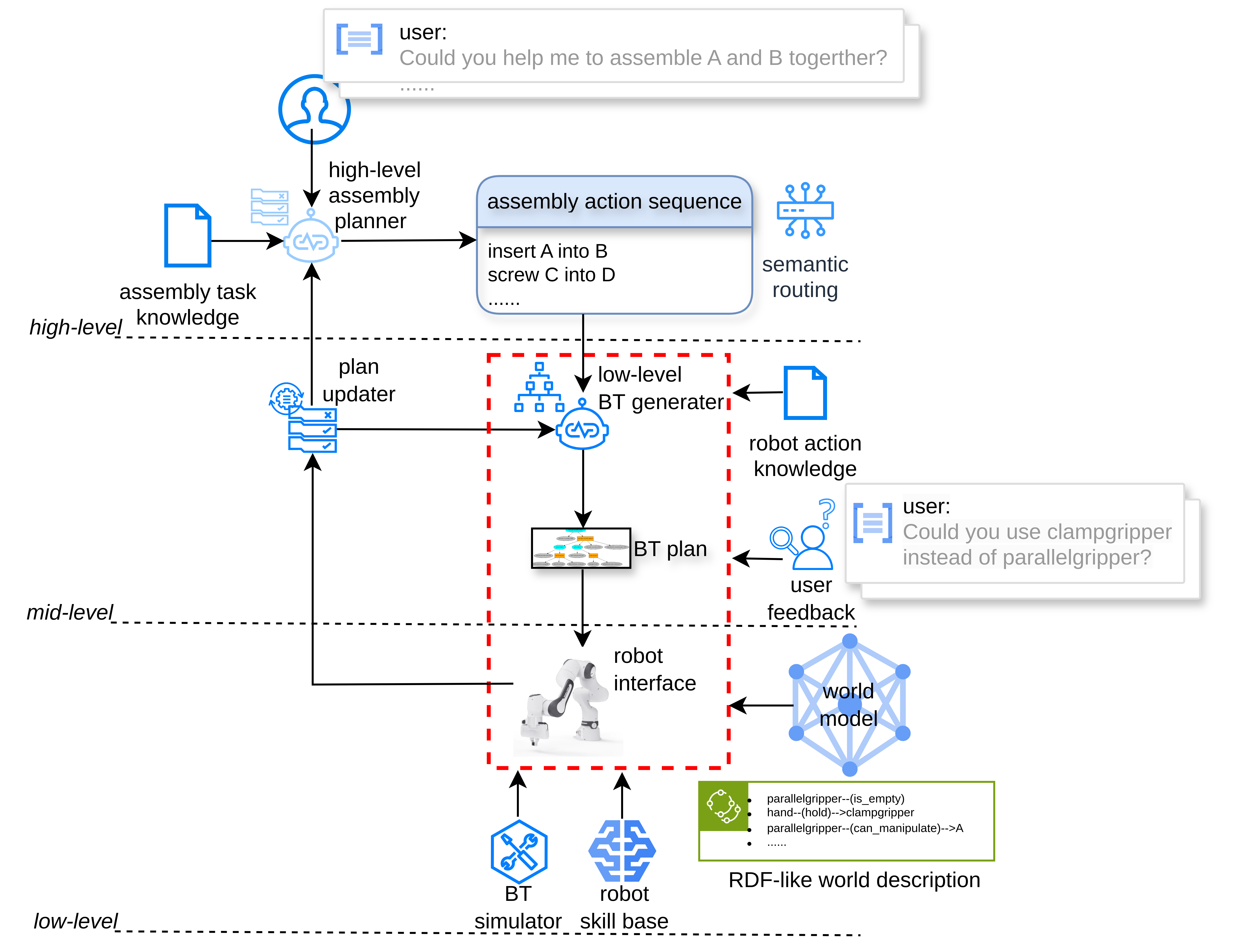}
    \caption{The basic workflow}\label{subfig:basic}
  \end{subfigure}%
  \hfill 
  \begin{subfigure}[b]{0.45\textwidth}
    \centering
    \begin{subfigure}[b]{0.4\textwidth}
      \includegraphics[width=\textwidth]{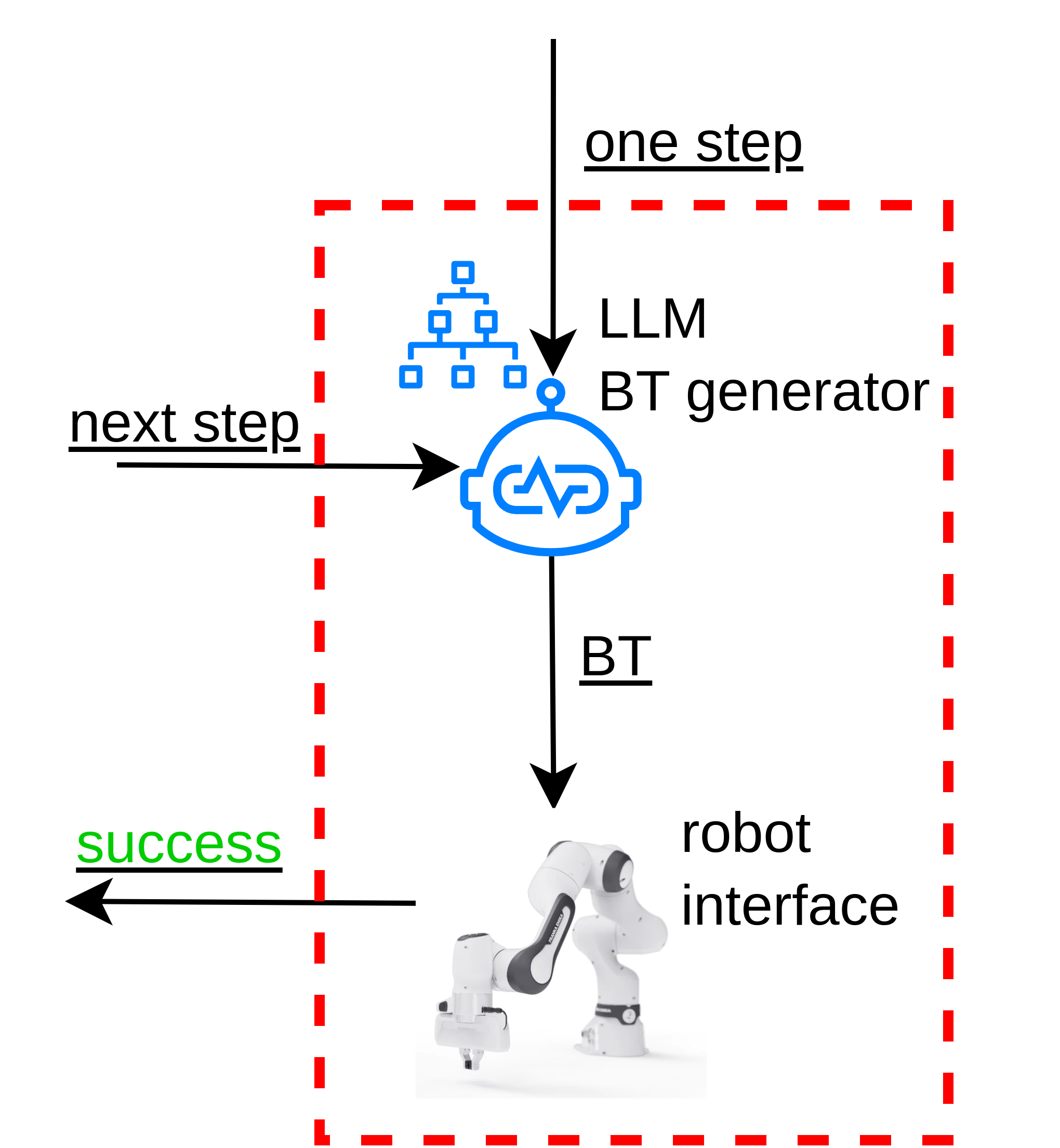}
      \caption{One-step generation}\label{subfig:onestep}
    \end{subfigure}%
    \hfill 
    \begin{subfigure}[b]{0.6\textwidth}
      \includegraphics[width=\textwidth]{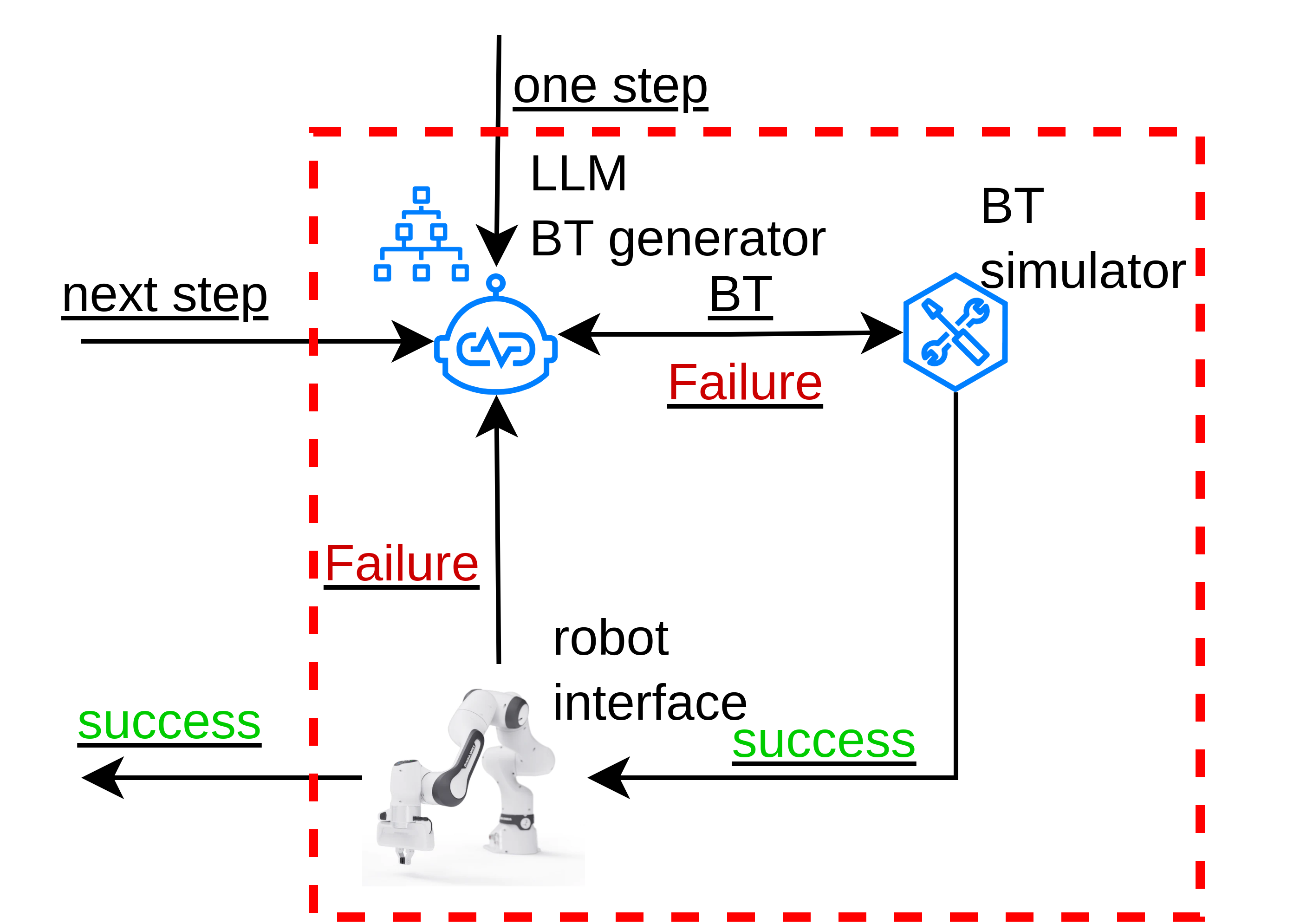}
      \caption{Iterative generation}\label{subfig:iter}
    \end{subfigure}\\[1ex] 
    
    \begin{subfigure}[b]{0.55\textwidth}
      \includegraphics[width=\textwidth]{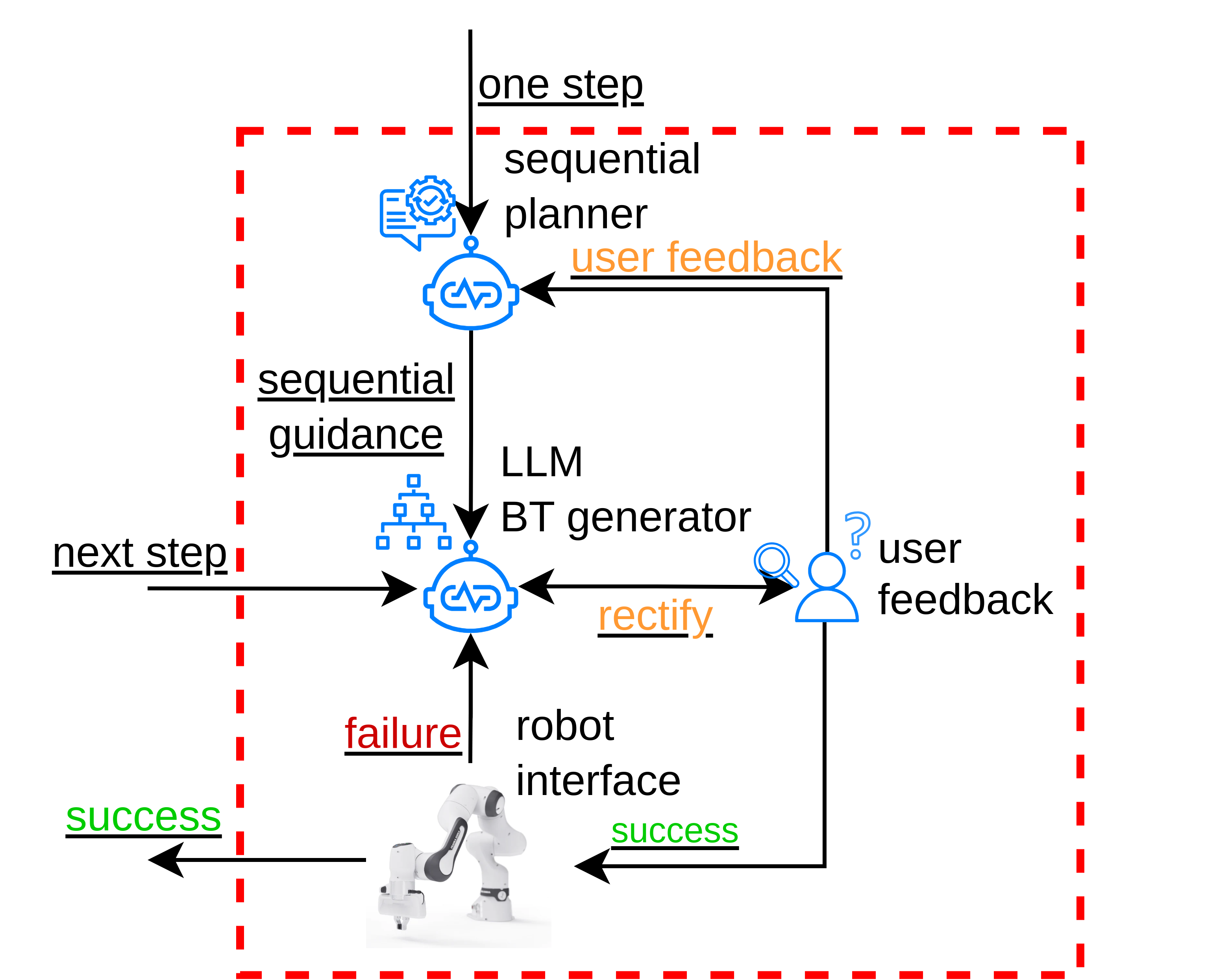}
      \caption{Human-in-the-loop generation}\label{subfig:hil}
    \end{subfigure}%
    \hfill 
    \begin{subfigure}[b]{0.45\textwidth}
      \includegraphics[width=\textwidth]{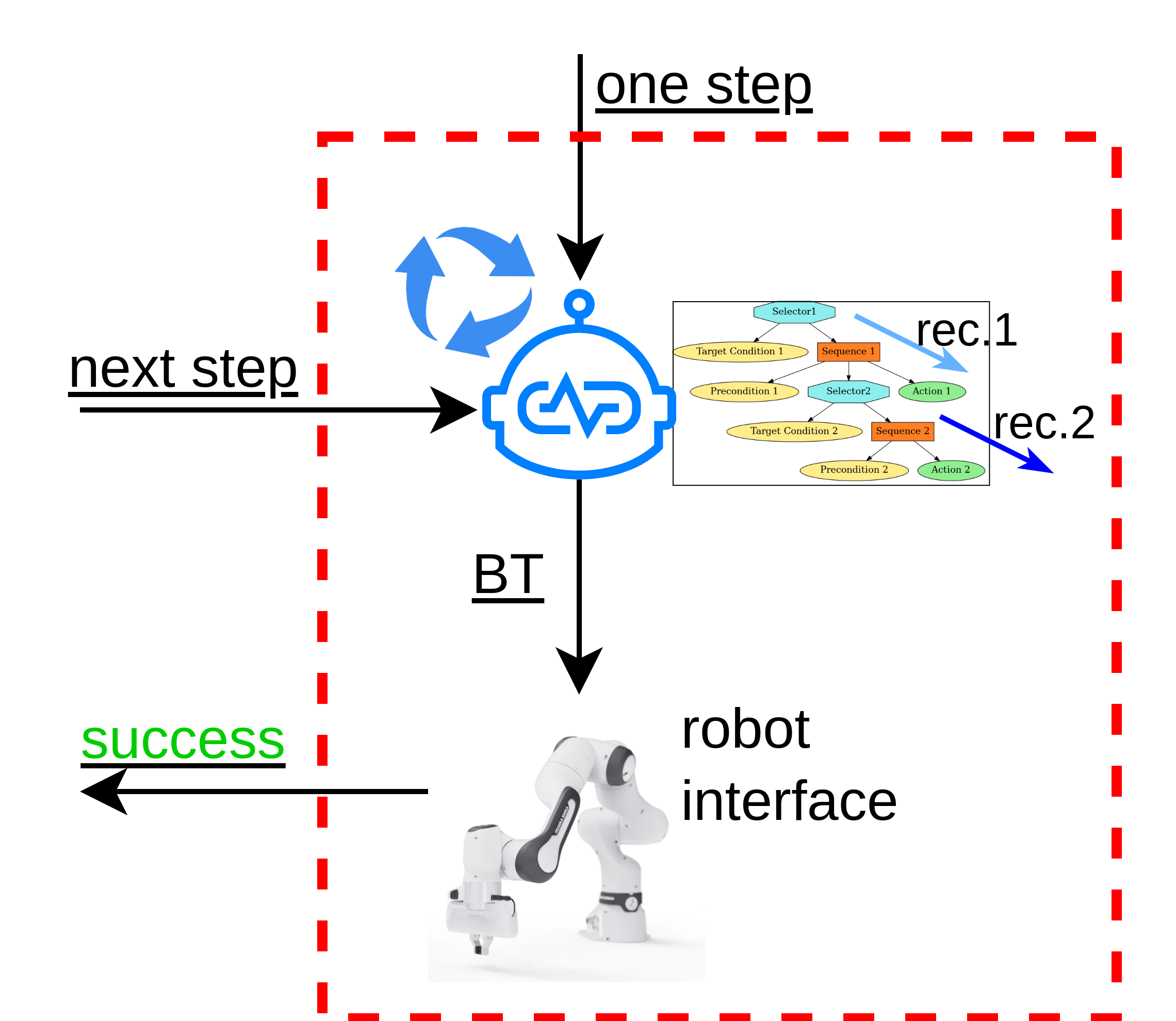}
      \caption{Recursive generation}\label{subfig:rec}
    \end{subfigure}
  \end{subfigure}

  \caption{The basic workflow and the four proposed methods, where the red dashed rectangles show the place in the workflow that the contents from different methods can substitute.} \label{fig:four methods}
\end{figure*}

This section presents our proposed human-guided LLM-based BT generation framework, as illustrated in Figure~\ref{subfig:basic}. It encapsulates four different methods for BT generation, namely \il{\item one-step, \item iterative, \item human-in-the-loop, and \item recursive generation (Figure~\ref{fig:four methods}(b-e)).}

\subsection{LLM-ENHANCED ROBOT TASK PLANNING AND EXECUTION FRAMEWORK}




In this framework, the process is started by user instructions, which are then processed by a high-level LLM-based assembly planner for action sequence generation. In this step, the necessary assembly task knowledge is provided in natural language similar to the way in \cite{wangPlanandSolvePromptingImproving2023}. 
The generated action sequence is then parsed to a low-level LLM-based BT generator as a planning target.
During the BT generation process, the knowledge of robot actions and world predicates, written in a PDDL-like form with natural language explanations, is utilized. The world state is also an important reference for BT generation, which is represented in an RDF-like format as in \cite{singhProgPromptProgramGeneration2023,yonedaStatlerStateMaintainingLanguage2023}. 
An example of the generated BT is shown in Figure \ref{fig:BT}. The robot interface then loads the generated BT and executes it with the help of the world model, which provides not only the world state but also the spatial data of the objects in the environment. Between BT generation and execution, it is possible to involve human feedback to guide the refinement of BT planning. 
To mitigate execution risks, simulation feedback can help pre-adjust BT plans, while a skill base equips the executor with diverse robotic actions, ensuring the assembly process’s accuracy and adaptability. For user inputs, semantic routers are applied to guide the workflow and improve the HRI experience.

The roles of human users in this framework are twofold.
On the one hand, at the beginning of the whole pipeline, the human teacher instructs the action sequence generation, using natural language to specify task constraints and goals.
On the other hand, in the human-in-the-loop BT planning process, the human user provides feedback by observing the execution of generated BT and giving new instructions, which enables effective HRI to improve generation efficiency and accuracy.

\subsection{LLM-ENHANCED BEHAVIOR TREE GENERATION METHODS}

Based on the framework introduced above, four LLM-based BT generation methods are designed, as introduced in Scheme 1 - 4 below. 

\subsubsection*{\schemeref{onestep} - One-step generation (Fig.~\ref{subfig:onestep})}\manuallabel{scheme:onestep}{1}
As shown in the figure, this method makes use of an LLM-based BT generator to generate BTs for the assembly step coming from the upstream module. 
    
\subsubsection*{\schemeref{iterative} - Iterative generation (Fig.~\ref{subfig:iter})}\manuallabel{scheme:iterative}{2}
This method leverages the BT simulator to help rectify and regenerate the behavior tree. The results of the simulation execution are taken as feedback to regenerate the BTs iteratively.
    
\subsubsection*{\schemeref{human} - Human-in-the-loop generation (Fig.~\ref{subfig:hil})}\manuallabel{scheme:human}{3}
This method applies a sequential planner to generate an action sequence first, which is then used to help guide the generation of the BT plan. It also applies a user feedback step to provide feedback in natural language to improve the generated BT plan.
    
\subsubsection*{\schemeref{recursive} - Recursive generation (Fig.~\ref{subfig:rec})}\manuallabel{scheme:recursive}{4}
This method applies our proposed BT expanding algorithm with the help of LLMs, in a recursive way. The algorithm does a preorder traversal in the BT and generates BTs for the unsatisfied nodes recursively, which is similar to the algorithms proposed in \cite{colledanchiseBlendedReactivePlanning2019} and \cite{liAdaptiveBehaviorTrees2022}. This LLM-enabled recursive BT expansion algorithm is summarized in Algorithm \ref{algo}.

\section{Results}\label{sec:result}
The framework proposed above is tested on a gear set assembly use case from the Siemens Robotic Assembly Challenge. The experiment setup is shown in Figure \ref{fig:robot} in the Appendix. We use GPT-4 as the LLM for all generation tasks. An example of behavior tree generation and execution process is presented in Figure \ref{fig:exe}.
The experiments were tested on 17 tasks for each BT generation Scheme. Each task is a distinct case in terms of variations in the initial world state and goal state.



The preliminary evaluation of four proposed methods, as detailed in Table \ref{tab:2}, highlights their varying efficiency and effectiveness in generating BTs within the framework. The one-step generation method exhibits perfect BT executability and a success rate of $70.58\%$. The unsuccessful generations are due to the logical incoherence of the BTs.
By looking into the failure cases, most failures are due to insufficient tree depth and the lack of well-defined actions, which shows the limitation of this method. 
The iterative method does not show an advantage over one-step generation because all BTs generated by both methods in the test cases are executable.
The human-in-the-loop method demonstrates a significant improvement in logical coherence and executability because of the incorporation of precise user feedback, which however sacrifices the performance in generation duration. 
The recursive method, while ensuring high logical coherence and good executability, incurs the longest generation time, reflecting its thoroughness in distributing the generation task across multiple recursive LLM invokes. These results highlight the trade-offs between generation time, complexity, and accuracy among the methods. 
Currently, the human-in-the-loop approach stands out for its high success rate and balance against efficiency and token consumption. 
Recursive method, though consuming a huge time and tokens, shows an excellent ability to make generated BTs logically coherent, which may be more beneficial when using smaller fine-tuned LLM instead of GPT4. 

To further investigate the capability of smaller LLMs for BT generation, we fine-tuned two models, LlaMA2-13B-Chat and Mistral-7B, using data gathered via the dynamic BT expansion method outlined by \cite{colledanchiseBlendedReactivePlanning2019} and the outputs from the in-context learning process. 
After training with a learning rate of $1\times10^{-5}$ over two epochs, we observed notable improvements in terms of our proposed performance metrics in preliminary testing.
Future work will delve deeper into the four proposed methods, utilizing the collected precise data from the in-context learning study, to investigate LLMs' potential in creating multi-level nested structures for planning long-horizon manipulation tasks. 

\begin{table}[!htb]
\begin{threeparttable}
\caption{Comparison of BT generation results in terms of success rate, logical coherence, executability, generation duration and token consumption.}
\label{tab:2}
\setlength\tabcolsep{0pt} 

\begin{tabular*}{\columnwidth}{@{\extracolsep{\fill}} l ccccc}
\toprule
     Method &  
     \multicolumn{3}{c}{Accuracy} & GD(sec.)\tnote{d} & TC\tnote{e} \\
\cmidrule{2-4}
     & SR\tnote{a} & LC\tnote{b} & Exec\tnote{c} \\
\midrule
     One-step & 12/17 & 12/17 & 17/17 & 49.11 & 5074.96 \\
     Iterative & 12/17 & 12/17 & 17/17 & 48.52 & 7770.13\\
     Human-in-the-loop & 16/17 & 16/17 & 17/17 & 85.02 & 7483.34\\
     Recursive & 13/17 & 17/17 & 13/17 & 231.04 & 50229.96\\
\bottomrule
\end{tabular*}


\end{threeparttable}
\end{table}



\newpage
\section*{ACKNOWLEDGMENT}
The authors acknowl- edge the financial support by the Bavarian State Ministry for Economic Affairs, Regional Development and Energy (StMWi) for the Lighthouse Initiative KI.FABRIK (Phase 1: Infrastructure as well as the research and development program under grant no. DIK0249).

\bibliographystyle{IEEEtran}
\bibliography{lib.bib, additional_lib, extra_lib}

\section*{APPENDIX}
\subsection{Experiment Setup}
\begin{figure}[htb]
\centering
\includegraphics[width=0.6\linewidth]{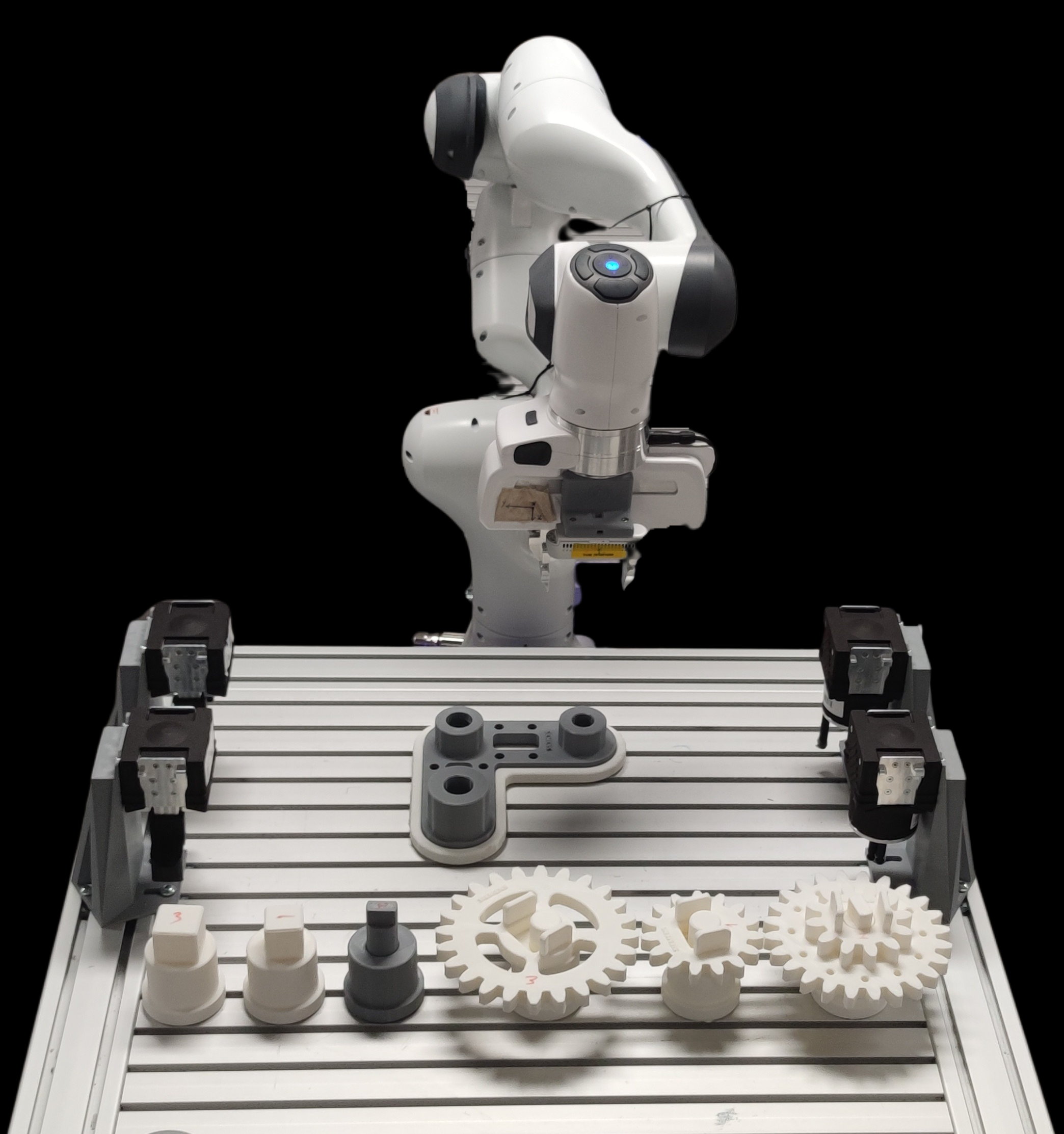}
\caption{Experiment setup with a franka panda robot, four toolcubes from Leverage, and a gearset from the Siemens Robot Assembly Challenge} \label{fig:robot}
\end{figure}

The experiment setup used in this work is shown in Fig.~\ref{fig:robot}. A single 7DOF manipulator, Franka Emika Panda, is controlled by joint torque commands via Franka Control Interface (FCI). The computed torques come from a Cartesian adaptive force-impedance controller, which tracks both desired force and motion simultaneously. To execute the actions in behavior trees, the action context is parsed to another custom control software to map the action to a pre-defined skill.

\subsection{Evaluation Metrics}
The metrics used for result evaluation are explained as follows:

\begin{description}
    \item[SR] Success Rate. A behavior tree can be taken as correct only if it is executable, semantically correct, and can achieve the goal state;  
    \item[LC] Logical Coherence. This means the execution order inside the BT aligns with its equivalent action sequence. Format errors are ignored here, e.g., two conditions in one condition node, which is not executable but semantically correct;
    \item[Exec] Executability. This means the BT follows the regulated format and can be executed. An incorrect BT plan is still considered executable, e.g., using a wrong tool, which is semantically incorrect but still executable;
    \item[GD] Generation Duration for generating an entire BT;
    \item[TC] Token Consumption for generating an entire BT.
\end{description}

\subsection{Exemplar Behavior Tree}
\begin{figure}[htb]
\centering
\includegraphics[width=\linewidth]{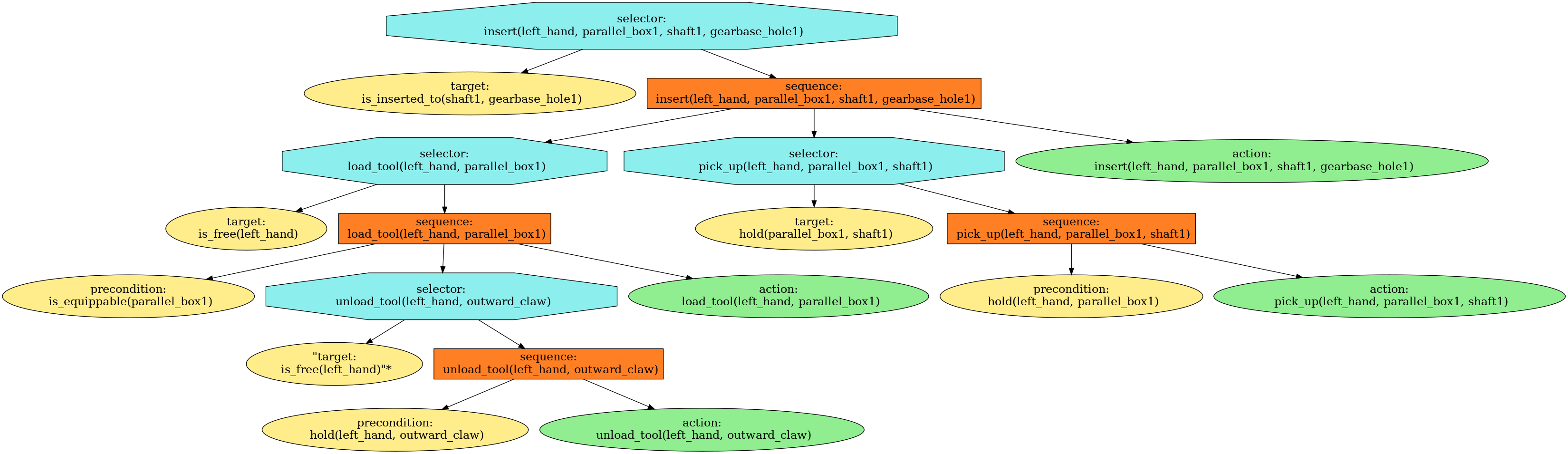}
\caption{An exemplar behavior tree generated from the assembly step "insert the shaft 1 into the gearbase hole 1".} \label{fig:BT}
\end{figure}
An exemplar BT generated from the assembly step "insert shaft1 into the gearbase hole1" is visualized in Fig.~\ref{fig:BT}. There are four types of nodes contained in this BT, namely, (1) \colorbox{YellowOrange}{Condition node}, (2) \colorbox{green}{Action node}, (3) \colorbox{SkyBlue}{Selector node}, (4) \colorbox{Bittersweet}{Sequence node}.

\subsection{Behavior Tree Expansion Algorithm}

\begin{figure*}[!htb]
\centering
\includegraphics[width=\linewidth]{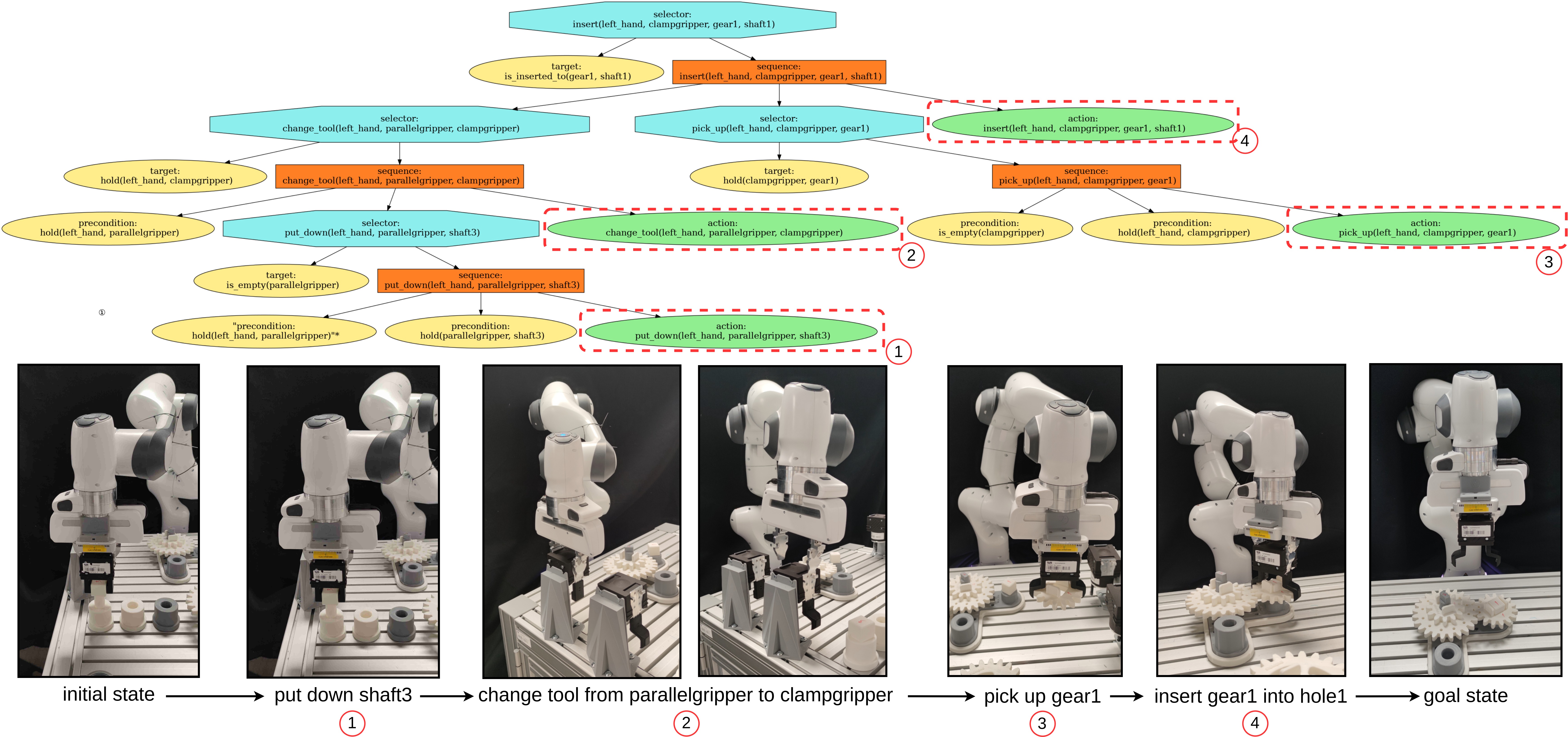}
\caption{\textbf{Robotic assembly of a gear set}. Shown are the generated behavior tree and the corresponding sequence of actions. The order of actions is labeled by number and shown from left to right, while their corresponding action nodes in the BT are colored green.} \label{fig:exe}
\end{figure*}

The detailed LLM-enabled Recursive Behavior-Tree Expansion algorithm is summarized in the table below.
Initially, the algorithm accepts an initial state, denoted as $s_0$, and a list of nodes, referred to as $node\_list$, to be expanded. The process begins with setting a temporary state, $s'_0$, equivalent to $s_0$, to maintain the current state across iterations. For each $node_i$ within the $node\_list$, the goal state $g_i$ is first determined associated with $node_i$ through the $\text{GetGoal}(node_i)$ function. Following this, a plan $plan_i$ that aims at achieving $g_i$ from the current state $s'_{i-1}$ is formulated using the $\text{MakePlan}(s'_{i-1}, g_i)$ function. This plan is represented as an action sequence, indicating the necessary steps to achieve the goal and its internal order. When a feasible plan is found by the MakePlan function, the algorithm estimates the resultant state $s'_i$ after executing $plan_i$ from $s'_{i-1}$ with the help of the $\text{EstimateState}(s'_{i-1}, plan_i)$ function. After that, the last action in $plan_i$, denoted as $a_i$, is utilized to generate a new tree, $tree_i$, through the $\text{MakeTree}(a_i)$ function. This new tree assists in identifying conditional child nodes $new\_node\_list$ via the $\text{GetCondChildren}(tree_i)$ function. The algorithm then recursively calls itself with $new\_node\_list$ and $s'_i$ to further expand the node structure. In scenarios where $plan_i$ is found to be empty, indicating the absence of a direct path to reach $g_i$ from the current state, the state remains unaltered ($s'_i = s'_{i-1}$), and the algorithm progresses to the next node in the list without any expansion for the current node. The algorithm stops when all the nodes in the $new\_node\_list$ are fulfilled by the initial state $s_0$.

\begin{algorithm}[!htb]
\caption{Behavior Tree Expansion Algorithm} \label{algo}
\begin{algorithmic}[1]

\Function{ExpandBehaviorTree}{$node\_list, s\_0$}
    \State $s'\_0 \gets s\_0$
    \For{each $node_i$ in $node\_list$}
        \State $g_i \gets$ \Call{GetGoal}{$node_i$}
        \State $plan_i \gets$ \Call{MakePlan}{$s'\_{i-1}, g_i$}
        \If{$\text{len}(plan_i) > 0$}
            \State $s'\_i \gets$ \Call{EstimateState}{$s'\_{i-1}, plan_i$}
            \State $a_i \gets plan_i[-1]$
            \State $tree_i \gets$ \Call{MakeTree}{$a_i$}
            \State $new\_node\_list \gets$ \Call{GetCondChildren}{$tree_i$}
            \State \Call{ExpandBehaviorTree}{$new\_node\_list, s'\_i$}
        \Else
            \State $s'\_i \gets s'\_{i-1}$
        \EndIf
    \EndFor
\EndFunction

\end{algorithmic}
\end{algorithm}

\subsection{Example of Behavior Tree Generation and Execution}

The execution process of a more complex BT is shown in Figure \ref{fig:exe}. The generated BT can be planned by any of the four proposed methods, which is planned to satisfy the upstream assembly target \textit{insert gear1 into shaft1} and represents its equivalent action sequence:
\begin{enumerate}
    \item \textbf{put\_down(left\_hand), parallelgripper, shaft3)}
    \item \textbf{change\_tool(left\_hand, parallelgripper, clampgripper)}
    \item \textbf{pick\_up(left\_hand, clampgripper, gear1)}
    \item \textbf{insert(left\_hand, clampgripper, gear1, shaft1)}
\end{enumerate}

In the initial state, the left hand is holding \textit{parallelgripper} and \textit{parallelgripper} is holding \textit{shaft 3}. According to the execution mechanism of BTs, action \textit{put\_down(left\_hand), parallelgripper, shaft3)} is executed first with its precondition nodes being satisfied by the initial state. After its execution, the condition node \textit{is\_empty(parallelgripper)} is satisfied, which allows the execution of action \textit{change\_tool(left\_hand, parallelgripper, clampgripper)} as its precondition node. This action changes the tool in the left hand from \textit{parallelgripper} to \textit{clampgripper}, fulfilling the condition node \textit{hold(left\_hand, clampgripper)}. This condition node serves as one of the precondition nodes of the action node \textit{pick\_up(left\_hand, clampgripper, gear1)}, which can then be ticked and executed. Finally, after the execution of all the actions mentioned above, the precondition nodes of the action \textit{insert(left\_hand, clampgripper, gear1, shaft1)}, namely \textit{hold(left\_hand, clampgripper)} and \textit{hold(clampgripper, gear1)}, are satisfied, allowing the execution of this final action. This action will fulfill the planning target of this BT, the target node \textit{is\_inserted\_to(gear1, shaft1)}. The BT returns \textit{SUCCESS} in the end, indicating a successful execution of the assembly task \textit{insert gear1 into shaft1}.




\end{document}